\documentclass[letterpaper]{article} % DO NOT CHANGE THIS

\usepackage{aaai24}  % DO NOT CHANGE THIS
\usepackage{times}  % DO NOT CHANGE THIS
\usepackage{helvet}  % DO NOT CHANGE THIS
\usepackage{courier}  % DO NOT CHANGE THIS
\usepackage[hyphens]{url}  % DO NOT CHANGE THIS
\usepackage{graphicx} % DO NOT CHANGE THIS
\usepackage{natbib}  % DO NOT CHANGE THIS AND DO NOT ADD ANY OPTIONS TO IT
\usepackage{caption} % DO NOT CHANGE THIS AND DO NOT ADD ANY OPTIONS TO IT
\usepackage{algorithm}
\usepackage{algorithmic}
\usepackage{amsmath}
\usepackage{amssymb}
\usepackage{multirow}
\usepackage{enumitem}
\usepackage[dvipsnames]{xcolor}
\newcommand*\samethanks[1][\value{footnote}]{\footnotemark[#1]}
\usepackage{multicol}  % 引入multicol包
\usepackage{cuted}
    
\usepackage{newfloat}
\usepackage{listings}
\DeclareCaptionStyle{ruled}{labelfont=normalfont,labelsep=colon,strut=off} % DO NOT CHANGE THIS
\floatstyle{ruled}
\newfloat{listing}{tb}{lst}{}
\floatname{listing}{Listing}
    \title{Scalable Geometric Fracture Assembly via Co-creation Space among Assemblers}

    \author {
        Ruiyuan Zhang\textsuperscript{\rm 1} \equalcontrib,\quad
        Jiaxiang Liu\textsuperscript{\rm 1} \equalcontrib,\quad
        Zexi Li\textsuperscript{\rm 1},\quad
        Hao Dong\textsuperscript{\rm 3},\quad
        Jie Fu\textsuperscript{\rm 2} \equalcontrib \thanks{Corresponding authors.},\quad
        Chao Wu\textsuperscript{\rm 1}  \samethanks
    }
    \affiliations {
        \textsuperscript{\rm 1} Zhejiang University\quad
        \textsuperscript{\rm 2} Hong Kong University of Science and Technology \quad
        \textsuperscript{\rm 3} Peking University \\
        \{zhangruiyuan, zjljx, zexi.li, chao.wu\}@zju.edu.cn\quad
        jiefu@ust.hk\quad
        hao.dong@pku.edu.cn
    }

\begin{document}
    \maketitle
    \begin{abstract}
        Geometric fracture assembly presents a challenging practical task in archaeology and 3D computer vision. Previous methods have focused solely on assembling fragments based on semantic information, which has limited the quantity of objects that can be effectively assembled. Therefore, there is a need to develop a scalable framework for geometric fracture assembly without relying on semantic information.
        To improve the effectiveness of assembling geometric fractures without semantic information, we propose a co-creation space comprising several assemblers capable of gradually and unambiguously assembling fractures. Additionally, we introduce a novel loss function, i.e., the geometric-based collision loss, to address collision issues during the fracture assembly process and enhance the results.
        Our framework exhibits better performance on both PartNet and Breaking Bad datasets compared to existing state-of-the-art frameworks. Extensive experiments and quantitative comparisons demonstrate the effectiveness of our proposed framework, which features linear computational complexity, enhanced abstraction, and improved generalization. Our code is publicly available at https://github.com/Ruiyuan-Zhang/CCS.
    \end{abstract}

\section{Introduction}

    Fracture assembly aims to reconstruct a broken object by composing its fractures. Manually assembling these fragments is time-consuming and requires precision. The task is complicated due to the vast number of potential combinations and the lack of clear instructions. Therefore, geometric fracture assembly is a practical but challenging task.

    \begin{figure}
        \centering
        \includegraphics[]{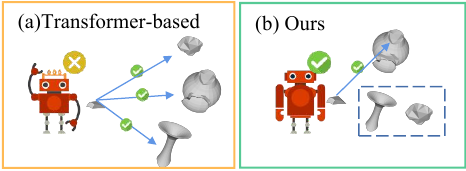} 
        \caption{
            Comparison of Transformer-based assembly (left) and our method (right). \textbf{Left:} Assemblers may become confused when presented with multiple fractures to assemble simultaneously, leading to an inability to assemble any parts. 
            \textbf{Right:} In our method, assemblers focus on essential fractures of parts, enabling efficient and effective task completion.
        }
        \label{fig_1}
    \end{figure}

    Previous studies have focused on tasks like archaeological fragment matching and 3D furniture assembly, using methods such as generative 3D part assembly with graph neural networks to understand part relationships \citep{wu2023leveraging, funkhouser2011learning, toler2010multi, zhang20223d, zhan2020generative, narayan2022rgl, lee2021ikea, li2020learning}. However, these approaches often face limitations in handling shapes with numerous fragments, also called scalability issue or multi-parts issue. Futhermore, they rely heavily on semantic information like part segmentation and ground-truth annotations \citep{zhang20223d, zhan2020generative, narayan2022rgl, lee2021ikea, li2020learning}.
    To address part relationship constraints, \citep{zhang20223d} introduced a transformer-based framework, IET, which assigns positions to parts and uses self-attention for positional interactions. However, this method's computational complexity increases significantly with the number of parts due to the quadratic scaling of attention mechanisms \citep{goyal2021coordination}.

    In this paper, we introduce the Co-creation Space, a novel approach that enables assemblers to predict the 6-DoF pose of geometric parts or fractures. This method, illustrated in Figure \ref{fig_overview}, involves assemblers competing for write access to update a shared workspace, thereby predicting part poses \citep{baars1993cognitive, dehaene2021consciousness}. The task of geometric fracture assembly is divided into specialized tasks, such as identifying the core fracture and locating relevant ones. Assemblers collaborate in a shared workspace, which replaces the pairwise interactions found in traditional dot-product attention methods. This approach ensures global coherence and reduces computational complexity to a linear scale relative to the number of assemblers \citep{baars1993cognitive}.
    As depicted in Figure \ref{fig_1}, the Co-creation Space also acts as an information bottleneck, denoted as \( R_t \) in Eq \ref{eq:r_t}, limiting the capacity of information channels between specialists. This ensures that only essential information is written into the workspace. Fig.\ref{fig_1}(a): when the assembler considers multiple parts/fractures simultaneously, it discovers that multiple components can match each other (e.g., a chair seat can connect with both the chair legs and the chair back; a fragment can link with other fragments at multiple angles). As the volume of information increases, it would cause confusion issue as other baselines (Fig.\ref{fig_1}(b)). Just as drivers pay more attention to important traffic signals rather than the onboard music while driving, the assembler requires a focus.

    Moreover, the presence of identical or similar information in fractures during assembly can cause ambiguity and hinder escaping local optima. We address this by introducing a geometric-based collision loss, depicted in Figure \ref{coll_loss}. This loss function actively separates identical or similar fractures that share the same location, guiding them towards more appropriate positions.
    We carried out comprehensive experiments on two major geometric fracture assembly datasets: PartNet \citep{mo2019partnet} and Breaking Bad \citep{sellan2022breaking}. Through numerous comparative experiments and ablation analyses, we compared our method with state-of-the-art works \citep{zhang20223d, zhan2020generative, narayan2022rgl} and verified the effectiveness of our proposed framework. Our method emphasizes scalability in assembly processes and can be integrated as a plug-and-play module to enhance geometric information extraction research.

\section{Related Works}
    \subsection{Geometric Shape Assembly}
        Geometric shape assembly involves combining multiple shapes to create a target object \citep{zhang20223d, zhan2020generative, narayan2022rgl,grason2016perspective,chen2022neural,funkhouser2011learning,jones2021automate,lee2021ikea,li2020learning, liu2023deep}, and it is an important problem in science and engineering \citep{funkhouser2011learning, zhang20223d, zhan2020generative, narayan2022rgl}. Previous research has focused on specific cases that simplify the problem, such as using identical fragments \citep{funkhouser2011learning,lee2022learning} or textured fragments\citep{leefragment}. However, in practical applications, the shapes and number of fractures can be arbitrary, requiring more general methods.

        Building on PartNet \citep{mo2019partnet}, studies like \citep{zhan2020generative, narayan2022rgl} have proposed graph-based learning methods for predicting 6-DoF poses of each part and assembling a shape. Similarly, \citep{zhang20223d} use a transformer-encoder to understand part relationships. However, these methods often rely heavily on semantic information of object parts, such as instance encoding \citep{zhang20223d}, instance label \citep{zhan2020generative}, and order information \citep{narayan2022rgl}, and become less effective with more parts. Recently, \citep{sellan2021breaking} introduced the Breaking Bad dataset, highlighting the challenge of assembling non-semantic fractures into complete shapes. This shows that fractured shape assembly is an ongoing issue. In response, we propose a new framework to tackle the scalability challenge in geometric shape assembly, demonstrating good performance on non-semantic datasets.

    \subsection{Shared Global Workspace}
        In cognitive neuroscience, the Global Workspace Theory (SGW) \citep{baars1993cognitive, dehaene2021consciousness} has been proposed to suggest an architecture allowing specialist modules to interact through a shared representation called workspace, a bandwidth-limited communication channel. The advantage of this approach is that it enables global coordination and coherence among different components, beyond just local or pairwise interactions.  Workspace can be modified by any specialist, and that is broadcast to all specialists. 
        \citep{goyal2021coordination} explore the use of such a communication channel in the context of deep learning, leading to greater abstraction and better generalization, which will be effective in large geometric fracture assembly tasks without semantic information. Based on this theory, \citep{liu2022stateful} propose a Centralized Training Decentralized Execution learning approach called Stateful Active Facilitator that enables agents to work efficiently in high-coordination and high-heterogeneity environments. In this paper, co-creation is proposed so that the assemblers competing for write access can update the workspace.
        Looking at it from another perspective, the process in SGW is similar to an information bottleneck (IB) that distills and extracts crucial information for our conscious awareness.
        In this paper, IB corresponds to $R_t$, used to address scalability issues. 
    
\section{Method}
    \begin{figure*}
        \centering
        \includegraphics[width=1.0\textwidth]{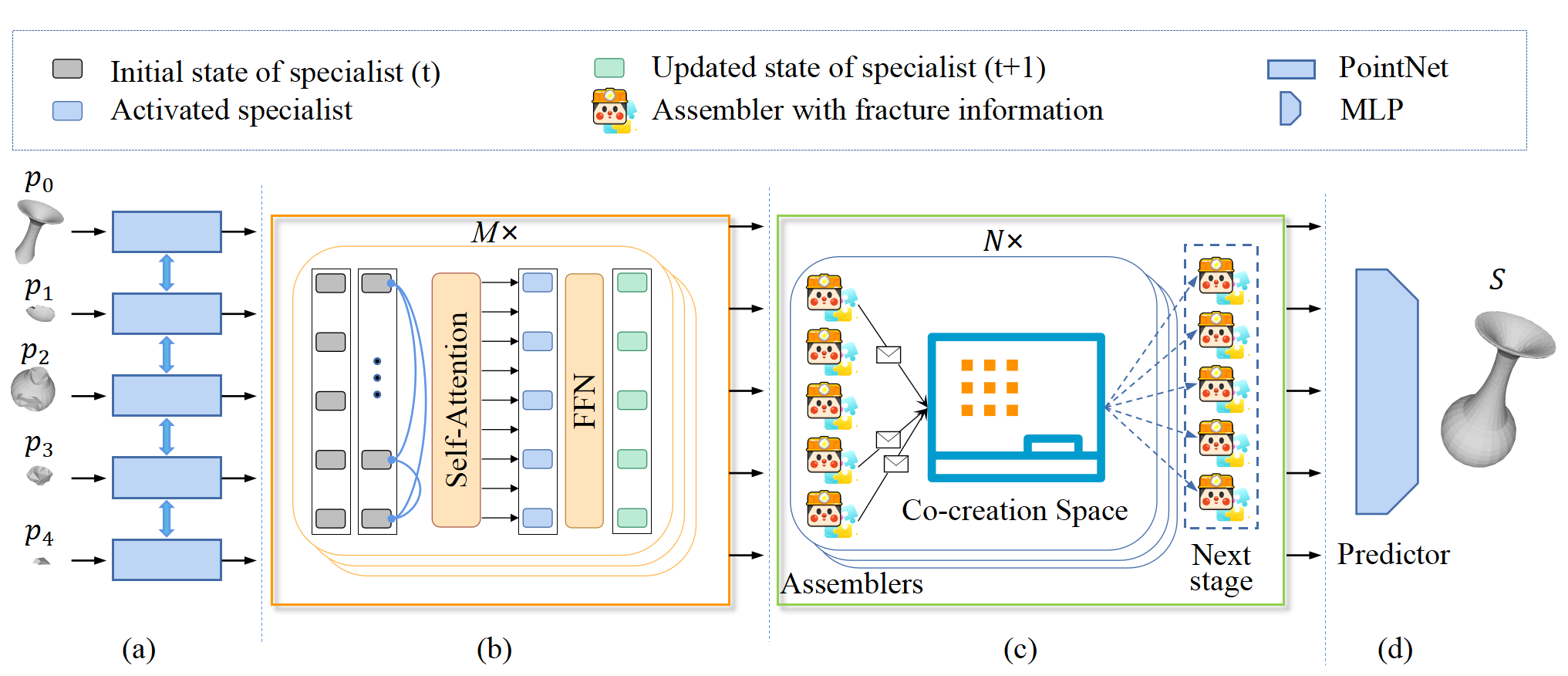} 
        \caption{The overall architecture of our framework. (a) geometric fractures feature extraction using shared parameters PointNet \citep{qi2017pointnet}, (b) geometric fractures relational reasoning and information routing, (c) complete the assembly task among assemblers using the co-creation space, (d) MLP predictor for pose estimation.}
        \label{fig_overview}
    \end{figure*}

    Let \(\mathcal{P} = \{p_i\}_{i=1}^N\) represent a set of geometric fracture point clouds, where \(p_i \in \mathbb{R}^{n_{pc} \times 3}\) and \(N\) denotes the number of parts which may vary for different 3D shapes. Our goal is to predict a set of 6-DoF fracture poses as \(Z = \{(R_i, T_i)\}_{i=1}^N\) in \(SE(3)\) space, where \(R_i \in \mathbb{R}^4\) and \(T_i \in \mathbb{R}^3\) denotes the rigid rotation and translation for each fracture, respectively. Then, we apply the predicted pose to transform the point cloud of each part and get the \(i\)-th fracture's predicted point cloud \(\mathcal{P}_i' = Z_i(p_i)\), in which \(Z_i\) is the joint transformation of \((R_i, T_i)\). And the complete shape can be assembled into \(S = \bigcup_{i=1}^N \mathcal{P}_i'\) as our predicted assembly result.
    In this work, we present a scalable geometric fractures assembly framework via a co-creation space among assemblers to assemble 3D shapes, which is illustrated in Figure \ref{fig_overview}.

    \subsection{Co-creation Space among Assemblers}
        In this section, we introduce co-creation space among assemblers, which serves as the core module of this framework. In this module at each computation stage indexed by $t$,  $n_a$ assemblers compete for write access to co-creation space, $n_a = N$. The contents of the co-creation space, in turn, are broadcast to all assemblers simultaneously.

        \paragraph{Step 1: Generating the message of assemblers.} 
        Each assembler $i$ receives a message containing geometric fractures' information $m_{i,t}$ at each time step $t$. 
        The initialization information is created through the routing function, which carries information about the current fracture and its relationship with other fractures. 
        The first step is external to the co-creation space, for details, please refer to \ref{rout_function}. 
        we denote the set of messages generated by the assemblers at time step $t$ by $M_t$:
        \begin{equation}
            M_t = \{m_{i,t} | 1 \leq i \leq N\}.
        \end{equation}

        \paragraph{Step 2: Writing into the co-creation space.} The message $M_t$ generated in step one is distilled into a latent state which we term as a \textbf{Co-creation Space}. The assemblers compete to write into the co-creation space, whose contents need to be updated in the context of new information received from different assemblers. This step ensures that only the critically important signals make it to the co-creation space, therefore preventing the co-creation space from being cluttered. 
        We represent the co-creation space state at time step $t$ by $R_t$. $R_t$ consists of $L$ slots $\{l_0, l_1, ..., l_{L-1}\}$, each of dimension $d_l$ so that $R_t \in \mathbb{R}^{L \times d_l}$.
        The messages in $M_t$ compete to write into each co-creation space's memory slot via a key-query-value attention mechanism. In this case, the query is a linear projection of the state of the current co-creation space memory content $R_t$, i.e., $\overset{\sim}{Q} = R_t \overset{\sim}{W^q}$, whereas the keys and values are linear projections of the messages $M_t$. Co-creation space state is updated as:
        \begin{equation} \label{eq:r_t}
            R_t = {\rm softmax}(\frac{\overset{\sim}{Q}(M_t\overset{\sim}{W^e})^T}{\sqrt{d_e}})M_t\overset{\sim}{W^v}.
        \end{equation}

        In this work, we use a top-$k$ softmax \citep{ke2018sparse} to select a fixed number of assemblers allowed to write in the co-creation space. 
        Similar to transformer \citep{vaswani2017attention}, we apply multiple heads to improve its expressive ability. 

        \paragraph{Step 3: Reading from the co-creation space.}  Each assembler then updates its state using the information broadcast from the co-creation space. We again utilize an attention mechanism to perform this operation. All the assemblers create queries $Q_t = \{q_{i,t} | 1 \leq i \leq N \} \in \mathbb{R}^{N \times d_e}$ where $q_{i,t} = W^q_{read}a_{i,t}$ and $a_{i,t}$ are encoded partial observations of one assembler. Generated queries are matched with the keys $K = R_tW^e \in \mathbb{R}^{l \times d_e}$ from the updated memory slots of co-creation space. As a result, the attention mechanism can be written as:
        \begin{equation}
            M_t' = {\rm softmax}(\frac{Q_tK^T}{\sqrt{d_e}})R_tW^v,
        \end{equation}
        where $M_t' = \{m_{i,t}' | 1 \leq i \leq N \}$. After receiving the broadcast information from the co-creation space, each assembler updates its state by a feedforward layer. This yield the new value $M_{t+1}$ for the $k$-th assembler, from which we start the next stage $(t+1)$.

        Co-creation is a shared workspace with limited capacity, which encourages specialization and compositionality among assemblers. IET \citep{zhang20223d} relies on pairwise interactions captured via an attention mechanism. Unfortunately, such attention mechanisms scale quadratically with the number of assemblers. Here, the computational complexity of the proposed method is linear in the number of assemblers.

    \subsection{Geometric Fractures Information Routing}
        \label{rout_function}

        \begin{figure*}
            \centering
            \includegraphics[width=0.9\textwidth]{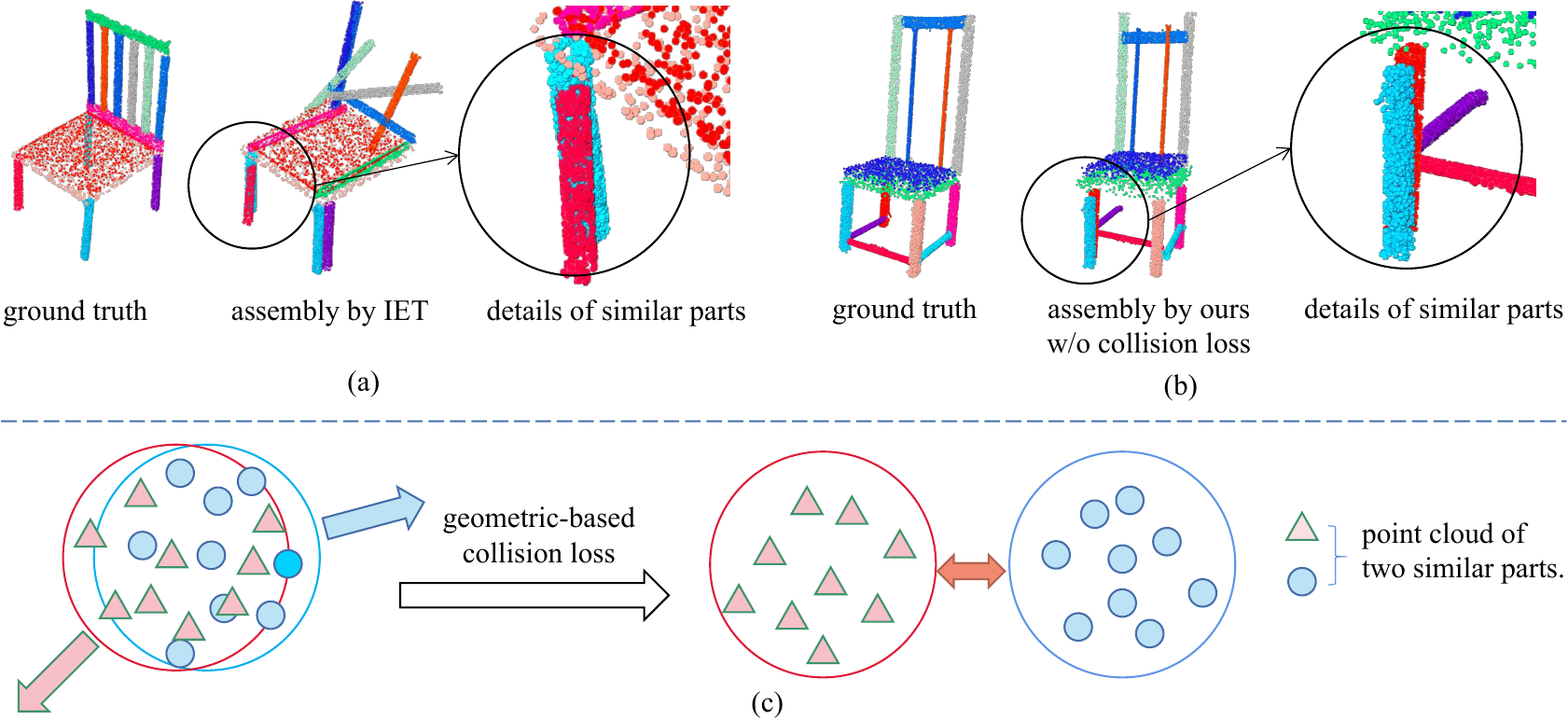}
            \caption{                                               
                The illustration of geometric-based collision loss (The red circles and the blue triangles represent 2 similar fractures).  \\
            }
            \label{coll_loss}
        \end{figure*}
        
        Inspired by modular deep learning \citep{pfeiffer2023modular}, before the parts assembly task, we employed an independent module to perform geometric information extraction and fractures relation reasoning, and route the aggregation information to the next module. A transformer-based architecture is recommended to learn the relationships between fractures. Transformer-encoder applies multiple self-attention layers\cite{zhu2023universal, zhu2023bridging} that aggregate information (e.g., geometry and posture) from the entire input sequence (here is a set of geometric fractures). The positional encoding is omitted as the input already contains the information about 3-dimensional XYZ coordinates. We didn't follow the standard formulation of the transformer. Instead, in order to keep the model architecture concise, this framework used the same Transformer structure as the co-creation space. Here the co-creation space can support information routing and relation reasoning at structure, but these two modules are different in terms of their functions in our framework. 
        
        The module discussed mainly performs information extraction and routing, while the co-creation space primarily completes the logical task of fracture assembly. Learning routing is a challenging research of modular deep learning, including \textit{training instability} \citep{pfeiffer2023modular}, \textit{module collapse} \citep{kirsch2018modular}, and \textit{overfitting} \citep{pfeiffer2023modular}. This is not the focus of this paper to be discussed, we will discuss the learned routing of Assembly in future work. We have provided several options to replace the geometric fracture information routing network structure, including ResNet \citep{he2016deep}, Transformer \citep{vaswani2017attention}, and GNN \citep{scarselli2008graph}.

    \subsection{Geometric-based Collision Loss}

        During assembling, fractures with similar or identical semantic information will be placed in the same location. As shown in Figure \ref{coll_loss}, other chair legs or similar parts are assembled in the same position. To solve this problem, previous methods manually input the semantic information of the parts, such as Instance Encoding \citep{zhang20223d}, Instance label \citep{zhan2020generative}, and order information \citep{narayan2022rgl}. However, these methods are not suitable for  assembly tasks that do not involve semantic information. Additionally, adding too much bias into the model will limit the generalization ability of these methods to new shape distributions.

        We hypothesize that the problem arises because of \textit{fractures may assemble at a local optimum point, which is extremely difficult to escape}. When one of the fractures is gently moved, the model will still pull the moved part back to its local optimum position during optimization. To verify our hypothesis, we propose a loss function called \textit{collision loss}, which warns and pushes away one of the fractures when placed in the same location. The displaced fracture will affect other losses, such as the shape chamfer distance loss, and gradually optimize it to a reasonable position. Experimental results show that our method effectively solves the ambiguity between fractures $c_i$, where $1 \leq i \leq N$. we define the Collision loss as:

        \begin{equation}
            \mathcal{L}_c = \frac{
                2 \times \sum^N_i \sum^N_{j < i} (1 - \log({C\left | \left | c_i - c_j \right | \right | _{2}})) \\
            }{N \times (N-1)},
        \end{equation}

        % I feel that this method can still be improved. From the experimental phenomena, it can be seen that if a relatively small C is used in the early stage and a larger C is used later, up to infinity, it seems to achieve the best effect.
        \noindent where $C$ is the hyperparameter of collision loss, that can be adjusted through grid search \cite{liu2023parameter}. Which presents a correlation with the distance to the cloud point distribution. \( \left\| c_i - c_j \right\|_2 \) represents the distance \( d \geq 0 \) between two parts or fractures. We only consider the loss \( l_{i,j} = 1 - \log(C \times d) \). The corresponding derivative is \( l_{i,j}' = - \frac{1}{d} \). As \( d \) increases, \( l_{i,j}' \) becomes larger, meaning its absolute value decreases because the derivative is negative. 
        The increase of the derivative leads to the slope of the $L_{i,j}$ becoming flatter.
        It indicates that the rate of decrease of the function slows down as \( d \) increases. 
        This characteristic explains that \( \mathcal{L}_c \) can timely separate two overlapping fractures without affecting the total loss of two non-overlapping parts. This can effectively solve the ambiguous issue. We add an ablation study for \( C \) with artifact dataset in Table \ref{table:grid_search}. 

    \subsection{Training Details}
        In the task of geometric fracture assembly, there are multiple possible solutions due to the interchangeable positions of the fractures and the ability to place decorative parts in different locations. In order to establish an uncertainty model and explore structural diversity, the Min-of-N (MoN) loss \citep{zhang20223d, zhan2020generative, narayan2022rgl}  and random noise $z_j \sim \mathcal{N}(0,1)$ were adopted. Here the overall framework was defined as $f$, while the ground truth pose was defined as $f^* $. 
        The MoN loss is used to calculate the error, which is defined as follows:
        \begin{equation}
            \mathcal{L}_{MoN} = \underset{1 \leq j \leq n }{\min} \mathcal{L}( f( \mathcal{P}, z_j ), f^*( \mathcal{P} )).
        \end{equation}
        Given a set of fracture point clouds $\mathcal{P}$, $f$ make $n$ predictions by perturbing the input with $n$ random vector $z_j$. Intuitively, it ensures at least one prediction as close as the ground-truth space. Following \citep{zhang20223d, zhan2020generative, narayan2022rgl}, we set $n$ = 5 in the experiment. The loss function, $\mathcal{L}$ is split into four categories similar to \citep{zhan2020generative}, collision loss was proposed by this work, for global and part-wise structural integrity.
    
        The translation is supervised by Euclidean loss $\mathcal{L}_t$, which measures the distance between the predicted translation $T_i$ and ground-truth translation $T_i^*$ for each fracture, 
        \begin{equation}
            \mathcal{L}_t = \sum^N_{i=1} \left | \left | T_i - T_i^* \right | \right | _{2} ^{2}.
        \end{equation}
    
        The rotation is supervised via Chamfer distance on the rotated fracture point cloud:
        \begin{equation}
        \begin{aligned}
            \mathcal{L}_{r} &= \sum_{i=1}^N\left( \sum_{x \in R_i (p_i)} \min _{\substack{y \in R_i^*(p_i)}}\|x-y\|_2^2  \right.\\ 
            &\left.  + \sum_{x \in R_i^*(p_i)} \min _{\substack{y \in R_i(p_i)}}\|x-y\|_2^2 \right) ,
        \end{aligned}
        \end{equation}
        in which the $R_i (p_i)$ and $R_i^*(p_i)$ repesent the rotated fracture points $p_i$ using the estimated rotation $R_i$ and the ground-truth $R^*_i$, respectively.

        To ensure comprehensive assembly quality, we have employed Chamfer distance (CD) to monitor the entire shape assembly $S$.

        \begin{equation}
            \mathcal{L}_s=\sum_{x \in S} \min _{y \in S^*}\|x-y\|_2^2+\sum_{y \in S^*} \min _{x \in S}\|x-y\|_2^2,
        \end{equation}
        
        \noindent where $S$ is the assembled shape and $S^*$ denotes the ground-truth. The total loss is defined as follows:
        \begin{equation} \label{eq:total_loss}
            \mathcal{L} = w_c \mathcal{L}_c + w_t \mathcal{L}_t + w_r \mathcal{L}_r + w_s \mathcal{L}_s,
        \end{equation}
        where $w_c$, $w_t$, $w_r$ and $w_s$ denote the weight of different losses, which are empirically determined.

\section{Experiments and Analysis}
    \begin{table*}
        \centering
        \caption{Quantitative evaluation on PartNet (Chair) and Breaking Bad datasets (Everyday and Artifact).}
        \label{results}
        \resizebox{\linewidth}{!}{
            \begin{tabular}{c|c|cccccc||cc} 
            \hline
            Metrics & Category & LSTM & Global & CM & DGL & RGL & IET & Ours & \(\triangle\) \\ 
            \hline
            \multirow{3}{*}{SCD (\(\times10^{-3}\))\(\downarrow\)} & Chair & 13.10 & 14.60 & 24.10 & 09.10 & 08.70 & 09.40 & \textbf{07.00} & \textcolor{YellowOrange}{-01.70} \\
            & Everyday & 20.42 & - & - & 15.12 & - & 14.29 & \textbf{14.26} & \textcolor{YellowOrange}{-00.03} \\
            & Artifact & 25.42 & - & - & 17.47 & - & 16.46 & \textbf{15.49} & \textcolor{YellowOrange}{-00.97} \\ 
            \hline
            \multirow{3}{*}{PA\(\uparrow\)} & Chair & 21.77 & 15.70 & 08.78 & 39.00 & 49.06 & 37.50 & \textbf{53.59} & \textcolor{YellowOrange}{+04.53} \\
            & Everyday & 18.30 & - & - & 26.40 & - & 26.94 & \textbf{28.57} & \textcolor{YellowOrange}{+01.63} \\
            & Artifact & 06.22 & - & - & 16.09 & - & 17.64 & \textbf{19.78} & \textcolor{YellowOrange}{+02.14} \\ 
            \hline
            CA\(\uparrow\) & Chair & 06.89 & 09.90 & 09.19 & 23.87 & 32.26 & 24.47 & \textbf{38.97} & \textcolor{YellowOrange}{+06.71} \\ 
            \hline
            \multirow{2}{*}{RMSE (R)\(\downarrow\)} & Everyday & 83.50 & - & - & 80.99 & - & 80.49 & \textbf{79.18} & \textcolor{YellowOrange}{-01.31} \\
            & Artifact & 86.20 & - & - & 82.64 & - & 79.94 & \textbf{77.59} & \textcolor{YellowOrange}{-02.35} \\ 
            \hline
            \multirow{2}{*}{RMSE (T) (\(\times10^{-2}\))\(\downarrow\)} & Everyday & 16.63 & - & - & 15.35 & - & \textbf{15.07} & 15.10 & \textcolor{YellowOrange}{+00.03} \\
            & Artifact & 17.50 & - & - & 16.05 & - & 15.72 & \textbf{15.50} & \textcolor{YellowOrange}{-00.22} \\
            \hline
            \end{tabular}
        }
    \end{table*}

    \begin{figure*}[t]
            \centering
            \includegraphics[width=\textwidth]{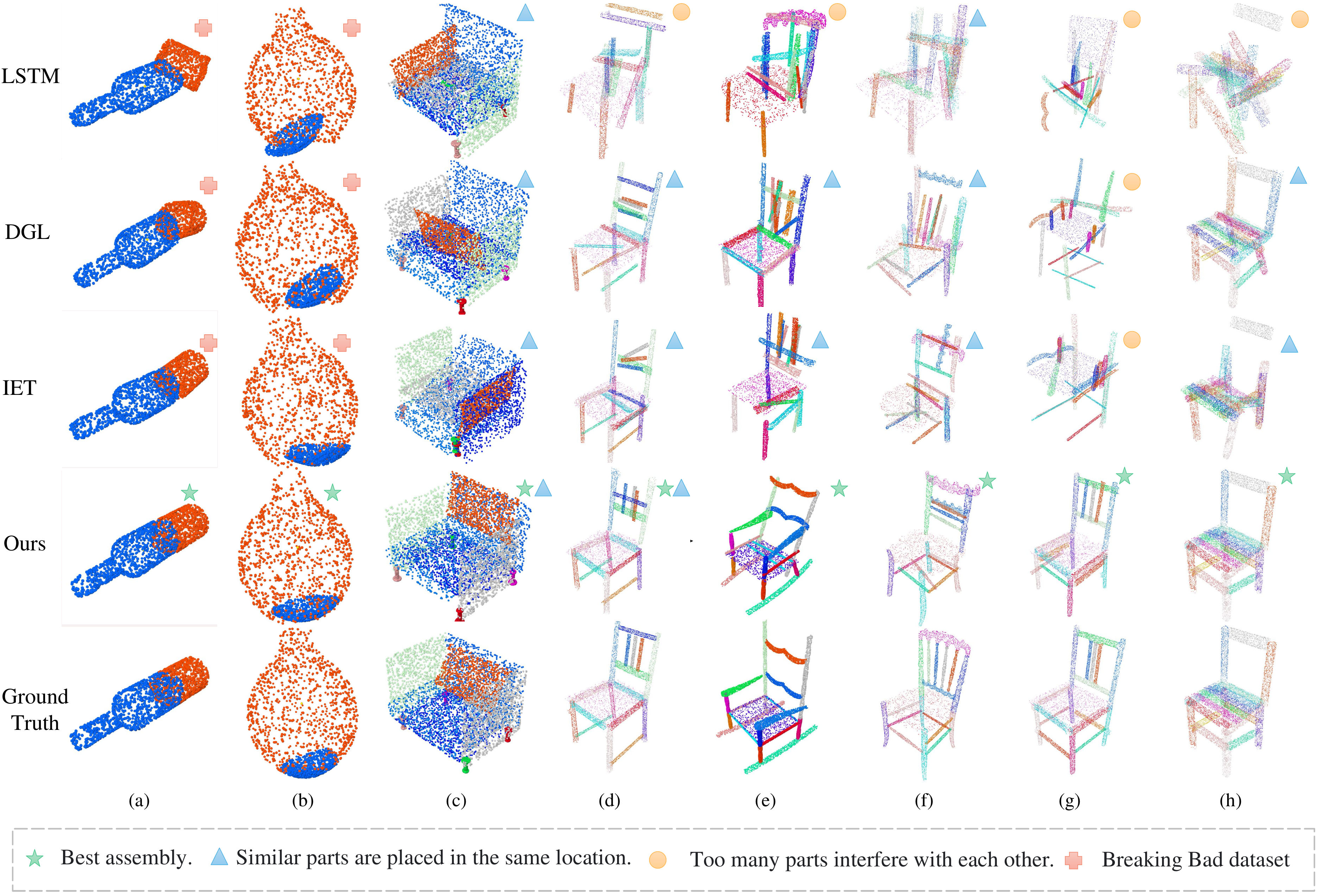}
            \caption{ 
                Visual comparisons between our algorithm and the baseline methods on Breaking Bad (a, b) and PartNet dataset (c, d, e, f, g and h). 
             }
            \label{fig_compare}
        \end{figure*}

\subsection{Dataset and Baselines}
        We evaluated our method and baselines on PartNet \citep{mo2019partnet} datasets and Breaking Bad \citep{sellan2022breaking, sellan2021breaking}. 
        PartNet is a large-scale shape dataset with fine-grained and hierarchical part segmentation. We utilize the Chair dataset with default train/validation/test being split in the dataset, which includes 6,323 chairs. 
        Breaking Bad dataset \citep{sellan2021breaking} contains a diverse set of shapes spanning everyday objects, artifacts without any manually annotated semantic information, e.g., instance label. It combines one million geometrically natural fracture patterns, which meets our needs. We used the categories of Everyday and Artifact. The number of parts or fractures used in all datasets ranges from 2 to 20.
        We compare the proposed method with Global \citep {li2020learning, schor2019componet}, LSTM \citep{wu2020pq}, CM \citep{sung2017complementme}, DGL \citep{zhan2020generative}, RGL \citep{narayan2022rgl} ,IET \citep{zhang20223d}. 
        To ensure a fair comparison, the implementations of other baselines are based on benchmark \citep{sellan2022breaking}.
        
    \subsection{Evaluation Metrics}
        The performance of our proposed method and baselines are measured by generating a variety of shapes and finding the closest shape to the ground truth using minimum matching distance \citep{achlioptas2018learning}. 
        To ensure a thorough evaluation, we use the metrics of part accuracy (PA) \citep{li2020learning}, connectivity accuracy (CA) \citep{zhan2020generative}, and shape Chamfer distance (SCD) \citep{zhan2020generative}, as employed by \citep{zhan2020generative}. PA and CA assess the precision of each individual part and the quality of the connections between them, respectively, while SCD evaluates the overall quality of the assembled shape. In addition, we computed the root mean squared error RMSE(R) \citep{sellan2022breaking} and root mean squared error RMSE(T) \citep{sellan2022breaking} to evaluate both rotation and translation prediction when conducting experiments using the Breaking Bad dataset.
        
  \subsection{Experimental Results and Analysis}
        We conducted a performance evaluation of our proposed method and various baselines, as illustrated in Table \ref{results}. Our proposed method demonstrated superior performance in most columns, particularly in the part and connectivity accuracy metrics. The red numerical value represents the number of instances where the proposed method outperforms the second-best method, and the green numerical value signifies the extent by which the proposed method lags behind the best-performing method. 
        
        The visual outcomes depicted in Figure \ref{fig_compare} and Figure B.2, Figure B.3 provide empirical evidence that our proposed approach surpasses the baseline methods in generating meticulously organized geometries. In contrast, the baseline methods often fail to achieve satisfactory assembly results. Subjectively speaking, our framework shows results that are almost indistinguishable from the ground truth. 
        Figure \ref{fig_compare} [d-h] shows that our framework can perform well even with many parts. This is because our framework includes a co-creation space, which has an information bottleneck property that promotes the emergence of better assembler professional skills, resulting in better generalization and logical reasoning abilities for the network\citep{tishby2000information,tishby2015deep,wu2020graph, ahuja2021invariance, goyal2021coordination}. 
        From Figure \ref{fig_compare} [c-h], we can see that in the baseline, similar part problems cannot be handled well when collisions occur. However, this problem is resolved in our framework. The success of our framework relies on the proposed collision loss, which can identify local optimal points during model training and overcome this state, and lead better results.
        Figure \ref{fig_compare} [a-b] shows that our framework still performs well on Breaking Bad, demonstrating that our framework has better relational reasoning abilities than previous baselines, even in  assembly tasks without semantic information.
        
        \begin{figure}[t]
            \centering
            \includegraphics[width=0.32\textwidth]{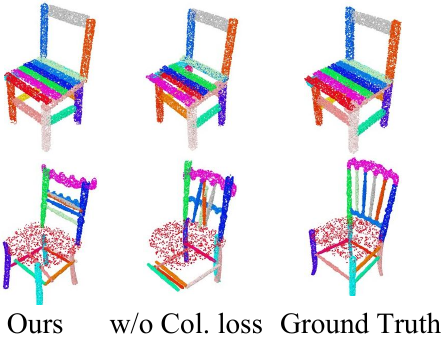}
            \caption{ 
                Collision loss ablation study visualization.
             }
             \label{fig_coll_ablation}
        \end{figure}

    \subsection{Ablation Study}
        \begin{table}[]
            \centering
            \small
            \renewcommand{\arraystretch}{0.8}
            \setlength\tabcolsep{4pt}
            \caption{The grid search of $w_c$ and $C$ in artifact.}
            \begin{tabular}{cc|cccc} 
                \hline
                $w_c$ & $C$ & SCD($\times 10^{-3}$)$\downarrow$ & PA$\uparrow$ & RSME(T)$\downarrow$ & RSME(R)$\downarrow$ \\ 
                \hline
                0.10   & 5     & 1.717                      & 19.53                      & 0.1618                     & 80.686                    \\
                0.10   & 10    & 1.610                      & 19.81                      & 0.1585                     & 81.049                    \\
                0.10   & 15    & 1.651                     & 19.10                      & 0.1589                     & 80.854                    \\
                0.10   & 20    & 1.615                     & 19.83                      & 0.1566                     & 80.054                    \\
                0.10   & 25    & 1.603                     & 19.83                      & 0.1560                     & {79.977}  \\
                0.10   & 30    & \textbf{1.581}            & {20.53}    & {0.1554}     & 80.811                    \\
                0.10   & 35    & 1.623                     & 19.39                      & 0.1562                     & 80.509                    \\
                0      & N/A   & 1.631                     & 19.81                      & 0.1579                     & 80.507                    \\
                0.05   & 30    & 1.685                     & 19.10                      & 0.1565                     & 81.028                    \\
                0.15   & 30    & {1.587}   & \textbf{20.54}             & \textbf{0.1553}            & 80.261                    \\
                0.20   & 30    & 1.645                     & 19.74                      & 0.1573                     & \textbf{79.837}           \\
                0.25   & 30    & 1.710                     & 19.38                      & 0.1561                     & 80.442                    \\
                \hline     
            \end{tabular}
            \label{table:grid_search}
        \end{table}   
        \begin{table}
            \centering
            \caption{The grid search of $k$ in our settings and ablation study. ``Col. loss'' denotes collision loss. \textcolor{blue}{Blue}/\textbf{bold} fonts highlight the suboptimal/best approach.}
            \begin{tabular}{c|ccc} 
            \hline
            Workspace & PA$\uparrow$   & SCD($\times10^{-3}$)$\downarrow$  & CA$\uparrow$    \\ 
            \hline
            w/o any workspace& 45.89 & 08.10 & 31.73  \\
            $k=1$ w/o Col. loss& 49.87 & 07.60 & 35.65  \\
            $k=5$ w/o Col. loss& 49.19 & 07.30 & 33.75  \\
            $k=10$ w/o Col. loss& \textcolor{blue}{52.53} & \textcolor{blue}{07.10} & \textcolor{blue}{38.40}  \\
            $k=15$ w/o Col. loss& 52.52 & 07.32 & 36.90  \\
            $k=20$ w/o Col. loss& 50.23 & 07.30 & 35.56  \\
            $k=10$ with Col. loss& \textbf{53.59} & \textbf{07.00} & \textbf{38.97}  \\
            \hline
            \end{tabular}
            % }
            \label{ablation_study}
        \end{table} 
    
    \paragraph{Co-creation Space Analysis}
        Co-creation space is a collaborative environment for all assemblers. In this context, the parameter $k$ represents the memory capacity of co-creation. This means that $k$ determines how much information each assembler can hold and process. 
        Table \ref{ablation_study} displays five different values of $k$, which range from low to high memory capacities. As the value of $k$ increases, assemblers can hold and process more information. However, there is a trade-off between memory capacity and communication efficiency among assemblers. 
        In this particular task, the optimal result is achieved when $k$=10. This means assemblers with a memory capacity of 10 can hold and process enough information to contribute effectively to the assembly process without becoming overwhelmed. 
        The visual assembly effect, as shown in Figure B.1, demonstrates the impact of memory capacity on the assembly process. At $k$=10, the assembly is best because co-creators can hold and process enough information to assemble the product correctly. However, when $k=x$ (where x is a lower or higher value than 10), problems such as misplaced parts or interference between similar parts can arise due to the information bottleneck.
        It is worth noting that our method failed to achieve a satisfactory result without co-creat space. Thus, it can be observed that our proposed co-creation space brought outstanding performance improvement consistent with qualitative analysis.

        \paragraph{Collision Loss}
             We carried out experiments to assess the effectiveness of the collision loss. As illustrated in Figure \ref{fig_coll_ablation}, we compared the performance of our method with and without the collision loss. The results show a significant enhancement in the model's capability to place similar parts in different locations with collision loss. This improvement is particularly noticeable when there are numerous comparable parts or a large number of parts. Thus, collision loss is effective to resolving ambiguity between similar parts during the fracture assembly procedure.
             Furthermore, we add an ablation study for \( w_c \) and $C$ on Table \ref{table:grid_search}. The values of \( w_t \), \( w_r \) and \( w_s \) follow previous works \cite{sellan2022breaking}.
        
        \paragraph{Coarse-to-fine}
            \begin{table}[]
                \caption{Coarse-to-fine ablation study.}
                \label{tb_c2f}
                \centering
                \begin{tabular}{c|ccc}
                \hline
                & PA\(\uparrow\) & SCD (\(\times10^{-3}\))\(\downarrow\) & CA\(\uparrow\) \\ 
                \hline
                IET & 37.50 & 09.40 & 24.47 \\
                Ours & 53.59 & 07.00 & 38.97 \\
                \(\triangle\) & \multicolumn{1}{c}{\textcolor{YellowOrange}{+16.09}} & \multicolumn{1}{c}{\textcolor{YellowOrange}{- 02.40}} & \multicolumn{1}{c}{\textcolor{YellowOrange}{+14.50}} \\ 
                \hline
                IET + CTF & 41.71 & 07.80 & 31.13 \\
                Ours + CTF & \textbf{55.92} & \textbf{06.20} & \textbf{42.69} \\
                \(\triangle\) & \multicolumn{1}{c}{\textcolor{YellowOrange}{+14.21}} & \multicolumn{1}{c}{\textcolor{YellowOrange}{- 01.60}} & \multicolumn{1}{c}{\textcolor{YellowOrange}{+11.56}} \\ 
                \hline
                \end{tabular}
            \end{table}

            Coarse-to-fine (CTF) is a sequential process that involves utilizing $x$ different networks to combine distinct parts into a complete shape\cite{zhan2020generative}, with each network focusing on more detailed information ($x=5$ in this paper). This approach enhances efficiency and accuracy by progressively refining the output at each step. The strategy was employed in both the DGL and IET. To verify its impact on experimental results, we implemented a version of the IET that employs CTF with five iterations. As depicted in Table \ref{tb_c2f}, our method still outperformed the others under this condition.
    
    \section{Conclusion}
        In this paper, we introduce a novel assembly paradigm that effectively addresses scalability issue without relying on semantic information. Firstly, we propose a co-creation space where assemblers compete for write access. This method facilitates step-by-step assembly, reducing confusion when dealing with multiple fractures simultaneously. Secondly, we have developed a unique geometric-based collision loss to minimize collision issues during assembly. Our method has been validated across all public datasets, and we are currently exploring its application in robotic hands for assembling real objects.

    \section{Acknowledgments}
        This work is supported by the National Key Research and Development Project of China (2021ZD0110505), National Natural Science Foundation of China (U19B2042), the Zhejiang Provincial Key Research and Development Project (2023C01043), University Synergy Innovation Program of Anhui Province (GXXT-2021-004), Academy Of Social Governance Zhejiang University, Fundamental Research Funds for the Central Universities (226-2022-00064).
        Furthermore, we thank the anonymous reviewers for their valuable comments and suggestions.
    
    \bibliography{aaai24}

% 附录文件
    \clearpage 
    
    \twocolumn[
        \begin{@twocolumnfalse}
            \section*{\centering{Supplementary Material for \emph{Co-creation Space Assembly}}}
            \section{A.Appendix}
            \subsection{Training Details}
                % We train Partnet and Breaking Bad in two different mad.
                For all the experiments. The learning rate (LR) is 1e-3. The epoch is 400 in our model and IET, 200 in LSTM and DGL. The warmup ratio is 0.05. The optimizer is Adam. The scheduler for LR is cosine. The LR decay factor is 100. The layer num was the same between IET and ours, e.g., 4. The batch size is 128 in Breaking Bad and 16 in PartNet. 
        
        \section{B.Appendix}
            This section will show more visual comparisons.
            
        \end{@twocolumnfalse}
    ]

    \begin{strip}
        \centering
        \includegraphics[width=0.9\textwidth]{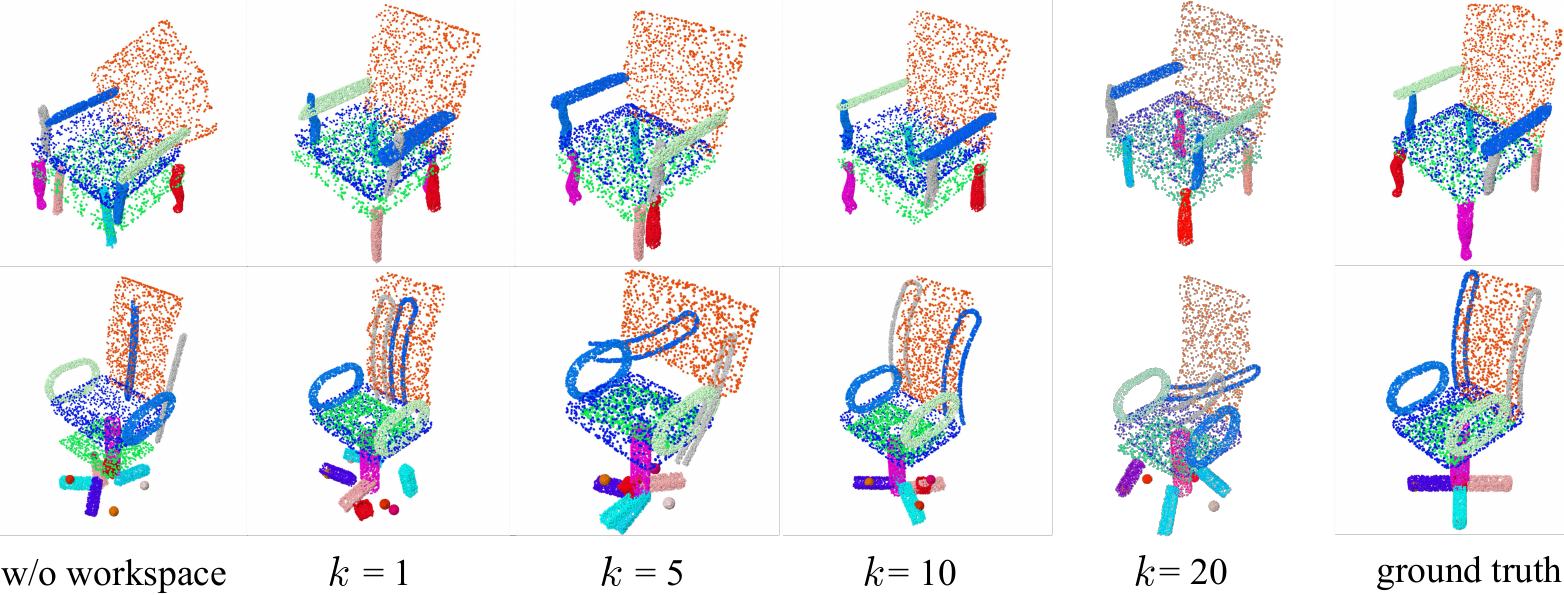}
        \captionof{figure}{The visualization of grid search for $k$.}
    \end{strip}

    \clearpage
    \begin{figure*}
        \label{Comparisons with baselines in PartNet}
        \centering
        % \fbox{\rule[-.5cm]{0cm}{4cm} \rule[-.5cm]{4cm}{0cm}}
        % \captionsetup{position=bottom}
        % \hspace{-3.5cm} 
        % \vspace{1cm}
        % \vspace{-1cm}
        \includegraphics[width=0.7\textwidth]{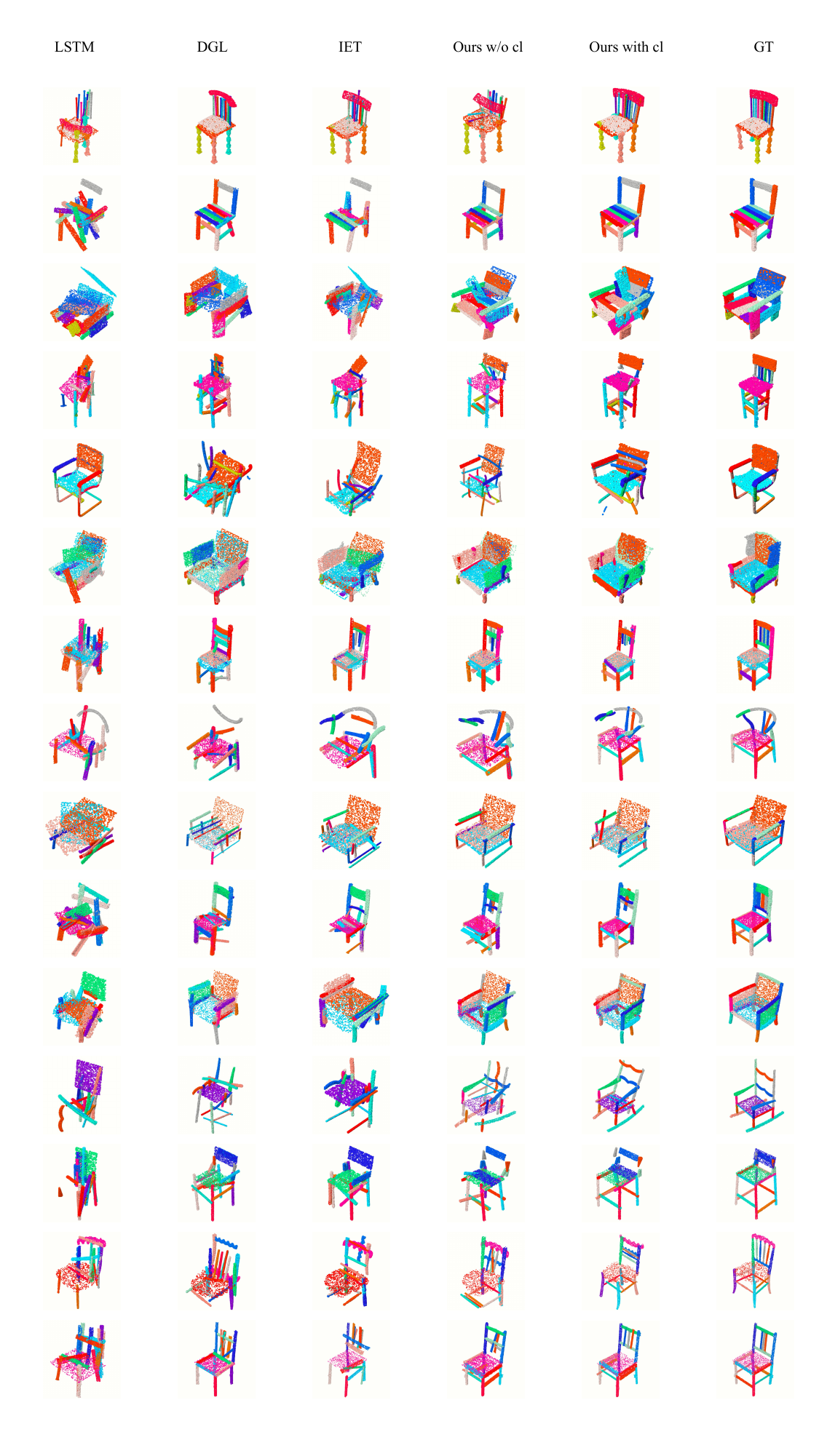} 
        \caption{more visual comparisons in PartNet.}
        % \label{fig_overview}
    \end{figure*}

    \begin{figure*}
        \centering
        \label{Comparisons with baselines in Breaking Bad}
        % \fbox{\rule[-.5cm]{0cm}{4cm} \rule[-.5cm]{4cm}{0cm}}
        % \captionsetup{position=bottom}
        % \hspace{-3.5cm} 
        % \vspace{1cm}
        % \vspace{-1cm}
        \includegraphics[width=0.7\textwidth]{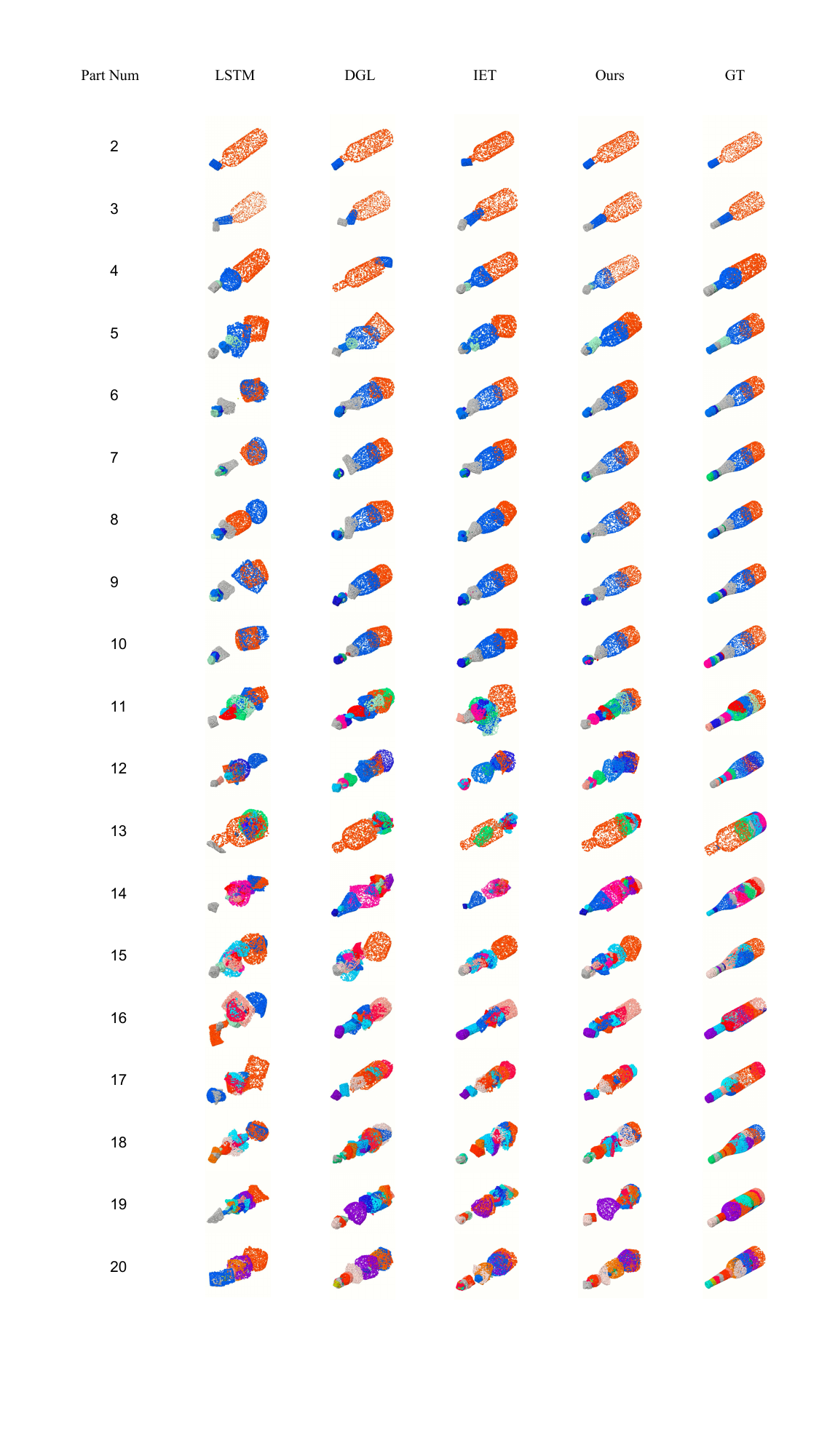} 
        \caption{more visual comparisons in Breaking Bad.}
        % \label{fig_overview}
    \end{figure*}
    % \end{multicols}

\end{document}